\documentclass[journal,twoside]{IEEEtran}
\usepackage{cite}
\ifCLASSINFOpdf
\else
  \usepackage[dvips]{graphicx}
\fi
\usepackage[cmex10]{amsmath}
\usepackage{array}
\ifCLASSOPTIONcompsoc
 \usepackage[caption=false,font=normalsize,labelfont=sf,textfont=sf]{subfig}
\else
 \usepackage[caption=false,font=footnotesize]{subfig}
\fi
\usepackage{fixltx2e}
\usepackage{url}
\usepackage{psfrag}

\usepackage{amssymb}

\usepackage{flushend}

\newcommand{\qm}[1]{``#1''}

\newcommand{\0}{\mathbf{0}}

\newcommand{\A}{\mathbf{A}}
\newcommand{\B}{\mathbf{B}}
\newcommand{\C}{\mathbf{C}}
\renewcommand{\th}{\boldsymbol\theta}
\newcommand{\m}{\boldsymbol\mu}

\newcommand{\G}{\boldsymbol\Gamma}
\newcommand{\balpha}{\boldsymbol\alpha}
\renewcommand{\r}{\mathbf{r}}

\renewcommand{\c}{\mathbf{c}}
\renewcommand{\v}{\mathbf{v}}
\newcommand{\x}{\mathbf{x}}
\newcommand{\y}{\mathbf{y}}
\newcommand{\h}{\mathbf{h}}
\renewcommand{\a}{\mathbf{a}}

\renewcommand{\b}{\mathbf{b}}

\renewcommand{\r}{\mathbf{r}}
\newcommand{\z}{\mathbf{z}}

\newcommand{\w}{\mathbf{w}}
\newcommand{\W}{\mathbf{W}}
\newcommand{\NN}{\mathcal{N}}
\newcommand{\p}{\mathring{p}}
\newcommand{\q}{\widetilde{q}}
\newcommand{\argmax}{\operatornamewithlimits{argmax}}

\newcommand{\esp}{\hspace{-2.5mm}}

\newcommand{\ww}{\mathbf{b}}

\newcommand{\eq}[1]{(\ref{#1})}

\usepackage[acronym]{glossaries}
\usepackage{xspace}

\newacronym{asr}{ASR}{Automatic Speech Recognition}
\newcommand{\asr}{\gls{asr}\xspace}

\newacronym{rtf}{RTF}{Real Time Factor}

\newacronym{wer}{WER}{Word Error Rate}

\newacronym{kkt}{KKT}{Karush-Kuhn-Tucker}

\newacronym{mmse}{MMSE}{Minimum Mean Square Error}
\newcommand{\mmse}{\gls{mmse}\xspace}

\newacronym{mse}{MSE}{Mean Square Error}

\newacronym{map}{MAP}{Maximum A Posteriori}
\newcommand{\map}{\gls{map}\xspace}

\newacronym{ml}{ML}{Maximum Likelihood}

\newacronym{cmllr}{CMLLR}{Constrained Maximum Likelihood Linear Regression}
\newcommand{\cmllr}{\gls{cmllr}\xspace}

\newacronym{mllr}{MLLR}{Maximum Likelihood Linear Regression}
\newcommand{\mllr}{\gls{mllr}\xspace}

\newacronym{sqp}{SQP}{Sequential Quadratic Programming}

\newacronym{qp}{QP}{Quadratic Program}

\newacronym{ff}{FF}{Frequency-Filtering}

\newacronym{mfcc}{MFCC}{Mel Frequency Cepstral Coefficient}
\newcommand{\mfcc}{\gls{mfcc}\xspace}

\newacronym{hmm}{HMM}{Hidden Markov Model}
\newcommand{\hmm}{\gls{hmm}\xspace}
\newcommand{\hmms}{\glspl{hmm}\xspace}

\newacronym{dct}{DCT}{Discrete Cosine Transform}

\newacronym{fft}{FFT}{Fast Fourier Transform}

\newacronym{dft}{DFT}{Discrete Fourier Transform}

\newacronym{remos}{REMOS}{Reverberation Modeling for Speech Recognition}
\newcommand{\remos}{\gls{remos}\xspace}

\newacronym{pdf}{pdf}{probability density function}
\newcommand{\pdf}{\gls{pdf}\xspace}

\newacronym{logmel}{logmelspec}{logarithmic melspectral}
\newcommand{\logmel}{\gls{logmel}\xspace}

\newacronym{mel}{melspec}{melspectral}

\newacronym{rvm}{RM}{Reverberation Model}

\newacronym{nm}{NM}{Noise Model}

\newacronym{rir}{RIR}{Room Impulse Response}

\newacronym{gmm}{GMM}{Gaussian Mixture Model}
\newcommand{\gmm}{\gls{gmm}\xspace}

\newacronym{dnn}{DNN}{Deep Neural Network}
\newcommand{\dnn}{\gls{dnn}\xspace}
\newcommand{\dnns}{\glspl{dnn}\xspace}

\newacronym{exmax}{EM}{Expectation Maximization}
\newcommand{\exmax}{\gls{exmax}\xspace}

\newacronym{rasta}{RASTA}{RelAtiveSpecTrA}

\newacronym{traps}{TRAPs}{TempoRAl Patterns}

\newacronym{cms}{CMS}{Cepstral Mean Subtraction}

\newacronym{cmn}{CMN}{Cepstral Mean Normalization}

\newacronym{3D}{3D}{three dimensional}

\newacronym{srr}{SRR}{Signal to Reverberation Ratio}

\newacronym{snr}{SNR}{Signal to Noise Ratio}

\newacronym{splice}{SPLICE}{Stereo Piecewise LInear Compensation for Environment}
\newcommand{\splice}{\gls{splice}\xspace}

\newacronym{pmc}{PMC}{Parallel Model Combination}
\newcommand{\pmc}{\gls{pmc}\xspace}

\newacronym{vts}{VTS}{Vector Taylor Series}
\newcommand{\vts}{\gls{vts}\xspace}

\newacronym{lms}{LMS}{Least Mean Square}

\newacronym{nlms}{NLMS}{Normalized Least Mean Square}

\newacronym{pnlms}{PNLMS}{Proportionate Normalized Least Mean Square}

\newacronym{rls}{RLS}{Recursive Least Squares}

\newacronym{lds}{LDS}{Linear Dynamical System}

\newacronym{klt}{KLT}{Karhunen-Loeve Transform}

\newacronym{lcmv}{LCMV}{Linearly Constrained Minimum Variance}

\newacronym{rwcp}{RWCP}{Real World Computing Partnership}

\newacronym{mint}{MINT}{Multiple input/output INverse Theorem}

\allowdisplaybreaks

\begin{document}
\title{A Bayesian Network View on\\Acoustic Model-Based Techniques for\\Robust Speech Recognition}

\author{Roland~Maas, Christian Huemmer,~\IEEEmembership{Student Member,~IEEE},
        Armin~Sehr,~\IEEEmembership{Member,~IEEE}, and~Walter~Kellermann,~\IEEEmembership{Fellow,~IEEE}%
\thanks{R. Maas, C. Huemmer, and W. Kellermann are with the institute
of  Multimedia Communications and Signal Processing, University of Erlangen-Nuremberg, 91058 Erlangen, Germany,
(e-mail: maas@LNT.de; huemmer@LNT.de; wk@LNT.de).}%
\thanks{A. Sehr is with the Department VII, Beuth University of Applied Sciences Berlin, 13353 Berlin, Germany,
(email: sehr@beuth-hochschule.de).}%
\thanks{The authors would like to thank the Deutsche Forschungsgemeinschaft (DFG) for supporting this work (contract number KE 890/4-2).}}%
\markboth{}%
{} %
\maketitle

\begin{abstract}
This article provides a unifying
Bayesian network view on various approaches for
acoustic model adaptation, missing feature, and uncertainty decoding that are well-known in the literature of robust automatic speech recognition.
The representatives of these classes can often be deduced from a Bayesian network that extends the conventional hidden Markov models used in speech recognition.
These extensions, in turn, can in many cases be motivated from an underlying observation model that relates
clean and distorted feature vectors. %
By converting the observation models into a Bayesian network representation, we formulate the corresponding compensation rules %
leading to a unified view on known derivations as well as to new formulations for certain approaches.
The generic Bayesian perspective provided in this contribution thus highlights structural differences and similarities between the analyzed approaches.
\end{abstract}

\IEEEpeerreviewmaketitle

\section{Introduction}

Robust \asr still represents a challenging research topic.
The main obstacle, namely the mismatch of test and training data,
can be tackled by enhancing the observed speech signals or features in order to meet the training conditions
or by compensating for the distorted test conditions in the acoustic model of the \asr system. 

Methods that modify the acoustic model are in general termed {\em (acoustic) model-based} or {\em model compensation} approaches and comprise inter alia the following sub-categories:
So-called {\em model adaptation} techniques mostly update the parameters of the acoustic model, i.e., of the \hmms, prior to the decoding of a set of observed feature vectors.
In contrast, {\em decoder-based} approaches re-adapt the \hmm parameters for each observed feature vector. The most common decoder-based approaches are {\em missing feature}
and {\em uncertainty decoding} that incorporate additional time-varying uncertainty information into the evaluation of the \hmms' probability density functions (pdfs).

Various model compensation %
techniques exhibit two (more or less) distinct steps that are taken interchangeably:
First, the compensation parameters need to be estimated and, second, the actual compensation rule is applied to the acoustic model.
The compensation rules can often be motivated based on an observation model that relates the clean and distorted feature vectors, e.g., in the \logmel or the \mfcc domain.

In this article, we show how the compensation rules can be deduced from the Bayesian network representations of the observation models for several
uncertainty decoding \cite{arrowood_2002, deng_dynamic_2005, droppo_uncertainty_2002, liao_uncertainty_2007, maas_formulation_2013},
missing feature \cite{cooke_robust_2001, raj_missing-feature_2005, kolossa_separation_2005, abdelaziz_decoding_2012},
and model adaptation techniques \cite{gales_model-based_1995, acero_hmm_2000, digalakis_speaker_1995, leggetter_maximum_1995, chien_linear_2003, wang_improving_2011,
hirsch_new_2008, raut_model_2006, sehr_frame-wise_2011, takiguchi_acoustic_2006}.
In addition, we give a Bayesian network description of the generic uncertainty decoding approach of \cite{ion_novel_2008}, of the \map adaptation technique \cite{gauvain_map_1992},
and of some alternative \hmm topologies \cite{ming_modelling_1996, maas_combined-order_2012}.
While Bayesian networks have been sporadically employed in this context before \cite{liao_uncertainty_2007, ion_novel_2008, abdelaziz_decoding_2012},
with this article, we give new formulations for certain of the considered algorithms in order to fill some gaps for a unified description.
Throughout the following, %
feature vectors are denoted by bold-face letters $\v_n$ with time index $n \in \{1,...,N\}$.
Feature vector sequences are written as \mbox{$\v_{1:N} = (\v_1,...,\v_N)$}.
Without distinguishing a random variable from its realization, a pdf over a random variable $z_n$ is denoted by $p(z_n)$.
For a normally distributed real-valued random vector $\z_n$ with mean vector $\m_{\z_n}$ %
and covariance matrix $\C_{\z_n}$, we write  \mbox{$\z_n \sim \mathcal{N}(\m_{\z_n}, \C_{\z_n})$} or 
\begin{equation}
	p(\z_n) = \mathcal{N}(\z_n ; \m_{\z_n}, \C_{\z_n}).
\end{equation}
To express that all random vectors of the set $\{\z_1,...,\z_N\}$ share the same statistics, we write $p(\z_n) = {\rm const.}$
or, in the Gaussian case,
\begin{equation}
	p(\z_n) = \mathcal{N}(\z_n ; \m_\z, \C_\z)
\end{equation}
with time-invariant mean vector $\m_\z$ and covariance matrix $\C_\z$.
Finally, for a Gaussian random vector $\z_n$ conditioned on another random vector $\w_n$, we write
\begin{equation}
	p(\z_n|\w_n) = \mathcal{N}(\z_n ; \m_{\z|\w_n}, \C_{\z|\w_n}),
\end{equation}
if the statistics of $\z_n$ depend only on time through $\w_n$,
i.e., if $\m_{\z|\w_n} = \m_{\z|\w_m}$ and $\C_{\z|\w_n} = \C_{\z|\w_m}$ for $\w_n = \w_m$ and $n,m \in \{1,...,N\}$.

The remainder of the article is organized as follows: After summarizing the employed Bayesian network view in Section~\ref{sec:bayes} and its difference to existing overview articles in Section~\ref{sec:merit},
this perspective is applied to uncertainty decoding, missing feature techniques, and other model-based approaches in Section~\ref{sec:examples}.
Finally, conclusions are drawn in Section~\ref{sec:conclusion}.

\section{A Bayesian Network View} \label{sec:bayes}

We start by reviewing the Bayesian network perspective on acoustic model-based techniques
that we use in Section~\ref{sec:examples} to compare different algorithms. %

\begin{figure}[!t]
	\centering
	\psfrag{A}[c][c]{$...$}
	\psfrag{q2}[l][l]{$q_{n\!-\!1}$}
	\psfrag{q3}[l][l]{$q_n$}
	\psfrag{x2}[l][l]{$\x_{n\!-\!1}$}
	\psfrag{x3}[l][l]{$\x_n$}
	\psfrag{z2}[l][l]{$\y_{n\!-\!1}$}
	\psfrag{z3}[l][l]{$\y_n$}
	\psfrag{y2}[l][l]{$\y_{n\!-\!1}$}
	\psfrag{y3}[l][l]{$\y_n$}
	\psfrag{c2}[l][l]{$\b_{n\!-\!1}$}
	\psfrag{c3}[l][l]{$\b_n$}
	\psfrag{B}[l][l]{(a)}
	\psfrag{C}[l][l]{(b)} \includegraphics[width=\columnwidth]{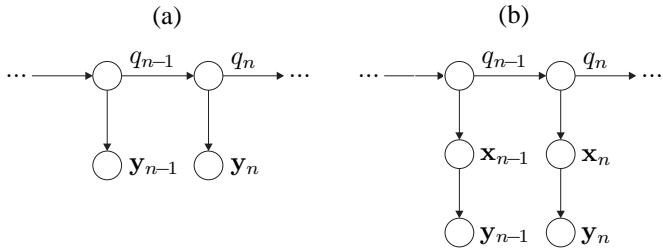}
	\caption{Bayesian network representation of (a) a conventional \hmm and (b) an \hmm incorporating latent feature vectors. Subfigure (b) is based on \cite{ion_novel_2008}.} %
	\label{fig:uncdec}
\end{figure}

Given a sequence of observed feature vectors $\y_{1:N}$, the acoustic score $p(\y_{1:N} | \W)$ of a sequence $\W$ of conventional \hmms,
as depicted in Figure~\ref{fig:uncdec}(a), is given by \cite{haeb_2011}
\begin{eqnarray}
	p(\y_{1:N} | \W) \esp&=&\esp \sum_{q_{1:N}} p(\y_{1:N},q_{1:N}) \label{eqn:hmm} \\
	\esp&=&\esp \sum_{q_{1:N}} \bigg\{\prod_{n=1}^N p(\y_n|q_n) \; p(q_n|q_{n-1})\bigg\}, \label{eqn:hmm_totalsplit}
\end{eqnarray}
where $p(q_1|q_0) = p(q_1)$. The summation goes over all possible state sequences $q_{1:N}$ through $\W$ superseding the explicit dependency on $\W$ at the right-hand side of (\ref{eqn:hmm}) and (\ref{eqn:hmm_totalsplit}).
Note that the pdf $p(\y_n|q_n)$ can be scaled by $p(\y_n)$ without influencing the discrimination capability of the acoustic score w.r.t. changing word sequences $\W$.
We thus define $\p(\y_n|q_n) = {p(\y_n|q_n)}/{p(\y_n)}$ for later use.

The compensation rules of a wide range of model adaptation, missing feature and uncertainty decoding approaches can be expressed by modifying the Bayesian network structure of a conventional \hmm
and applying the inference rules of Bayesian networks \cite{bishop_pattern_2006} -- potentially followed by suitable approximations to ensure mathematical tractability.
While some approaches postulate a certain Bayesian network structure, others indirectly define a modified Bayesian network by
assuming an observed feature vector $\y_n$ to be a {\em distorted} version
of an underlying {\em clean} feature vector $\x_n$, which is introduced as latent variable in the \hmm %
as, e.g., in Figure~\ref{fig:uncdec}(b).
In the latter case, the relation of $\y_n$ and $\x_n$ can be expressed by an analytical observation model $f(\cdot)$ that incorporates certain compensation parameters $\b_n$: %
\begin{equation}
	\y_n = f(\x_n, \b_n).
\end{equation}
Note that here it is not distinguished whether $\y_n$ is the output of a front-end enhancement process or a noisy or reverberant observation that is directly fed into the recognizer.
By converting the observation model to a Bayesian network representation,
the pdf $p(\y_n | q_n)$ in (\ref{eqn:hmm_totalsplit}) can be derived exploiting the inference rules of Bayesian networks \cite{bishop_pattern_2006}.
For the case of Figure~\ref{fig:uncdec}(b), the observation likelihood in (\ref{eqn:hmm_totalsplit}) would, e.g., become:
\begin{eqnarray}
	p(\y_n| q_n) \esp&=&\esp \int p(\x_n, \y_n|q_n) d\x_n \nonumber \\
	\esp&=&\esp \int p(\x_n|q_n) p(\y_n|\x_n) d\x_n,
\end{eqnarray}
where the actual functional form of $p(\y_n | \x_n)$ depends on the assumptions on $f(\cdot)$ and the statistics $p(\b_n)$ of $\b_n$.

The abstract perspective taken in this paper reveals a fundamental difference between model adaptation approaches on the one hand and missing feature and uncertainty decoding approaches on the other hand:
Model adaptation techniques usually assume $\b_n$ to have constant statistics over time \cite{liao_uncertainty_2007, kolossa_robust_2011}, i.e.,
\begin{equation}
	p(\b_n) = {\rm const.}, \mbox{ for } n \in \{1,...,N\}.
\end{equation}
or to be a deterministic parameter vector of value $\b$, i.e.,
\begin{equation}
	p(\b_n) = \delta(\b_n - \b),
\end{equation}
where $\delta(\cdot)$ denotes the Dirac distribution.
In contrast, missing feature and uncertainty decoding approaches typically assume $p(\b_n)$ to be a time-varying pdf \cite{liao_uncertainty_2007, kolossa_robust_2011}. %

As exemplified in Section \ref{sec:examples}, the Bayesian network view conveniently illustrates the underlying statistical dependencies %
of model-based approaches.
If two approaches share the same Bayesian network, their underlying joint pdfs over all involved random variables share the same decomposition properties.
However, some crucial aspects are not reflected by a Bayesian network: The particular functional form of the joint pdf, %
potential approximations to arrive at a tractable algorithm, %
as well as the estimation procedure for the compensation parameters. %
While some approaches estimate these parameters through an acoustic front-end, others derive them from clean or distorted data.
For clarity, we entirely focus in this article on the compensation rules while ignoring the parameter estimation step.
We also neglect approaches that apply a modified training method to conventional \hmms without exhibiting a distinct compensation step, %
as it is characteristic for, e.g., the case for discriminative \cite{heigold_discriminative_2012}, multi-condition \cite{matassoni_hidden_2002} or reverberant training \cite{sehr_novel_2010}.

\section{Merit of the Bayesian Network View} \label{sec:merit}

\renewcommand{\thesubsection}{\mbox{\Alph{subsection}}}
In the past decades, a range of survey papers and books have been published summarizing the state-of-the-art in noise and reverberation-robust \asr \cite{gong_speech_1995,lee_stochastic_1998,huo_robust_2001,droppo_environmental_2008,kolossa_robust_2011,yoshioka_making_2012,virtanen_techniques_2013}.
Recently, a comprehensive review of noise-robust \asr techniques was published in \cite{li_overview_2014} providing a taxonomy-oriented framework by distinguishing
whether, e.g., prior knowledge, uncertainty processing or an explicit distortion model is used or not.
In contrast to \cite{li_overview_2014} and previous survey articles, we pursue a threefold goal with this article:
\begin{itemize}
	\item First of all, we aim at classifying all considered techniques along the same dimension by motivating and describing them with the same Bayesian network formalism.
	Consequently, we do not conceptually distinguish whether a given method employs a time-varying pdf $p(\b_n)$,
	as in uncertainty decoding, or whether a distorted vector $\y_n$ is a preprocessed or a genuinely noisy or reverberant observation.
	Also the distinction of implicit and explicit observation models dissolves in our formalism.
	\item As a second goal, we aim at closing some gaps by presenting new derivations and formulations for some of the considered techniques.
	For instance, the Bayesian network representations of the concepts in
	Subsections \ref{ssec:splice}, \ref{ssec:jud}, \ref{ssec:ion}, \ref{ssec:cmllr}, \ref{ssec:mllr}, \ref{ssec:map}, \ref{ssec:bayes_mllr}, \ref{ssec:takiguchi}, and \ref{ssec:cond_comb_hmm}
	of Section \ref{sec:examples}
	have not been presented so far.
	Moreover, the links to the Bayesian network framework via the formulations in \eq{eqn:mi}, \eq{eqn:si}, \eq{eqn:cmllrDirac}, \eq{eqn:map1}, \eq{eqn:bayesMLLR2}, \eq{eqn:takiguchiPDF}
	are explicitly stated for the first time in this paper.
	\item The third goal of the Bayesian network description is to provide a graphical illustration that allows to easily overview a broad class of algorithms and
	to immediately identify their similarities and differences in terms of the underlying statistical assumptions.
\end{itemize}
By establishing new links between existing concepts, such an abstract overview should therefore also serve as a basis for revealing and exploring new directions.
Note, however, that the review presented in this paper does not claim to cover all relevant acoustic model-based techniques and is rather meant as an inspiration to other researchers.

\section{Examples} \label{sec:examples}

In the following, we consider the compensation rules of several acoustic model-based from a Bayesian network view.
The concepts in Subsections \mbox{\ref{sec:examples}-\ref{ssec:arrow}} to \mbox{\ref{sec:examples}-\ref{ssec:ion}} belong to the category of {\em uncertainty decoding} and
those in Subsections \mbox{\ref{sec:examples}-\ref{ssec:imputation}} to \mbox{\ref{sec:examples}-\ref{ssec:sig_dec}} to the class of {\em missing feature} approaches.
Furthermore, the methods summarized in Subsections \mbox{\ref{sec:examples}-\ref{ssec:pmc}} to \mbox{\ref{sec:examples}-\ref{ssec:takiguchi}} are commonly referred to as {\em acoustic model adaptation} techniques.
The concepts in Subsection \mbox{\ref{sec:examples}-\ref{ssec:cond_comb_hmm}} are examples of model-based techniques using modified \hmm topologies.

\begin{figure}[!t]
	\centering
	\psfrag{A}[c][c]{$...$}
	\psfrag{e2}[l][l]{(a)}
	\psfrag{e3}[l][l]{(b)}
	\psfrag{e4}[l][l]{(c)}
	\psfrag{e5}[l][l]{(d)}
	\psfrag{a2}[l][l]{$q_n$}
	\psfrag{k2}[l][l]{$k_n$}
	\psfrag{x2}[l][l]{$\x_n$}
	\psfrag{c2}[l][l]{$\y_n$}
	\psfrag{b2}[l][l]{$\b_n$}
	\psfrag{d2}[l][l]{$s_n$}
	\psfrag{a3}[l][l]{$\q_n$}
	\includegraphics[width=\columnwidth]{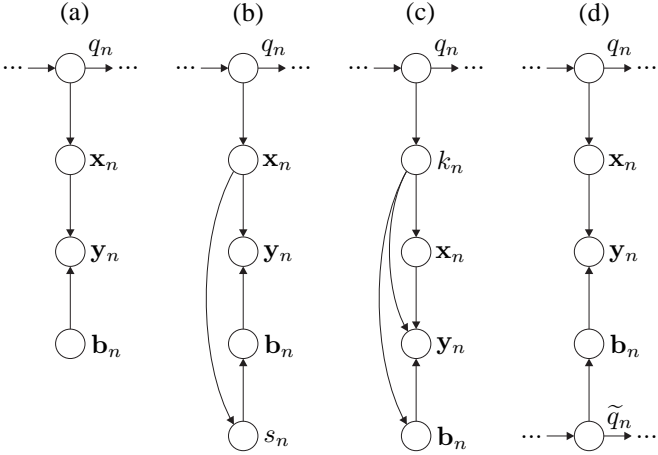}
	\caption{Bayesian network representation of different model compensation techniques. Detailed descriptions are given in the text. Subfigure (d) based on \cite{liao_uncertainty_2007}.}
	\label{fig:vertall}
\end{figure}

\subsection{General Example of Uncertainty Decoding} \label{ssec:arrow} %

A fundamental example of uncertainty decoding %
can, e.g., be extracted from \cite{holmes_using_1997, arrowood_2002, kristjansson_accounting_2002, deng_exploiting_2002, benitez_including_2004, stouten_model-based_2006, delcroix_static_2009}.
The underlying observation model can be identified as
\begin{eqnarray} \label{eqn:genericLinModel}
	\y_n = \x_n + \b_n \mbox{ with } \b_n \sim \NN(\0, \C_{\b_n}),
\end{eqnarray}
where $\y_n$ and $\C_{\ww_n}$ often play the role of an enhanced feature vector, e.g., from a Wiener filtering front-end \cite{benitez_including_2004},
and a measure of uncertainty from the enhancement process, respectively.
Thus, the point estimate $\y_n$ can be seen as being enriched by the additional reliability information $\C_{\ww_n}$.
The observation model is representable by the Bayesian network in Figure~\ref{fig:vertall}(a).
Exploiting the conditional independence properties of Bayesian networks \cite{bishop_pattern_2006},
the compensation of the observation likelihood in (\ref{eqn:hmm_totalsplit}) leads to
\begin{eqnarray}
	p(\y_n| q_n) \esp&=&\esp \int p(\x_n|q_n) p(\y_n|\x_n) d\x_n \nonumber \\
	\esp&=&\esp \int \NN(\x_n ; \m_{\x|q_n}, \C_{\x|q_n}) \, \NN(\y_n ; \x_n, \C_{\b_n})  d\x_n \nonumber \\ 
	\esp&=&\esp \NN(\y_n ; \m_{\x|q_n}, \C_{\x|q_n} + \C_{\b_n}). \label{eqn:arrowood}
\end{eqnarray}
Without loss of generality, a single Gaussian pdf $p(\x_n | q_n)$ is assumed
since, in the case of a \gmm, the linear mismatch function \eq{eqn:genericLinModel} can be applied to each Gaussian component separately.

\subsection{Dynamic Variance Compensation} \label{ssec:dvc}

The concept of dynamic variance compensation \cite{deng_dynamic_2005} is based on a reformulation of the log-sum observation model \cite{deng_exploiting_2002}:
\begin{eqnarray}
	\y_n = \x_n + \log(1+\exp(\widehat{\r}_n - \x_n)) + \b_n
\end{eqnarray}
with $\widehat{\r}_n$ being a noise estimate of any noise tracking algorithm and $\ww_n \sim \NN(\0, \C_{\ww_n})$ a residual error term.
Since the analytical derivation of $p(\y_n|q_n)$ is intractable, an approximate pdf is evaluated based on the assumption of $p(\x_n|\y_n)$ being Gaussian
and that the compensation %
can be applied to each Gaussian component of the \gmm separately \cite{deng_dynamic_2005} such that the observation likelihood in (\ref{eqn:hmm_totalsplit}) becomes
according to Figure~\ref{fig:vertall}(a):
\begin{eqnarray}
	\p(\y_n|q_n) \esp&=&\esp \int p(\x_n|q_n) \frac{p(\x_n|\y_n)}{p(\x_n)} d\x_n \nonumber \\
	\esp&\approx&\esp \int p(\x_n|q_n) p(\x_n|\y_n) d\x_n \nonumber \\
	\esp&\approx&\esp \int \NN(\x_n ; \m_{\x|q_n}, \C_{\x|q_n}) \nonumber \\
	&&\esp \NN(\x_n ; \m_{\x|\y_n}, \C_{\x|\y_n})  d\x_n \nonumber \\
	\esp&=&\esp \NN(\m_{\x|q_n}; \m_{\x|\y_n}, \C_{\x|q_n} + \C_{\x|\y_n}),
\end{eqnarray}
where the first approximation can be justified if $p(\x_n)$ is assumed to be significantly \qm{flatter}, i.e., of larger variance, than $p(\x_n|\y_n)$.
The estimation of the moments $\m_{\x|\y_n}$, $\C_{\x|\y_n}$ of $p(\x_n|\y_n)$ represents the core of \cite{deng_dynamic_2005}.

\subsection{Uncertainty Decoding with SPLICE} \label{ssec:splice}

The \splice approach, first introduced in \cite{deng_large_2000} and further developed in \cite{deng_high-performance_2001, li_deng_recursive_2003},
is a popular method for cepstral feature enhancement based on a mapping learned from stereo (i.e., clean and noisy) data \cite{li_overview_2014}.
While \splice can be used to derive an \mmse \cite{deng_high-performance_2001} or \map \cite{deng_large_2000} estimate that is fed into the recognizer,
it is also applicable in the context of uncertainty decoding \cite{droppo_uncertainty_2002}, which we focus on in the following.
In order to derive a Bayesian network representation of the uncertainty decoding version of \splice \cite{droppo_uncertainty_2002}, %
we first note from \cite{droppo_uncertainty_2002} that one fundamental assumption is:
\begin{eqnarray}
	p(\x_n|\y_n,s_n) = \NN(\x_n; \y_n + \r_{s_n}, \G_{s_n}),
\end{eqnarray}
where $s_n$ denotes a discrete region index. Exploiting the symmetry of the Gaussian pdf
\begin{eqnarray}
	p(\x_n|\y_n,s_n) \esp&=&\esp \NN(\x_n; \y_n + \r_{s_n}, \G_{s_n}) \nonumber\\
	\esp&=&\esp \NN(\y_n - \x_n; -\r_{s_n}, \G_{s_n})
\end{eqnarray}
and defining $\b_n = \y_n - \x_n$, we identify the observation model to be
\begin{equation}
	\y_n = \x_n + \b_n
\end{equation}
given a certain region index $s_n$.
In the general case of $s_n$ depending on $\x_n$, the observation model can be expressed by the Bayesian network in Figure~\ref{fig:vertall}(b) with %
\begin{eqnarray}
	p(\b_n | s_n) = \NN(\b_n; -\r_{s_n}, \G_{s_n}).
\end{eqnarray}
By introducing a separate prior model
\begin{eqnarray}
	p(\y_n) \esp&=&\esp \sum_{s_n} p(s_n) \, p(\y_n|s_n) \nonumber \\ 
	\esp&=&\esp \sum_{s_n} p(s_n) \, \NN(\y_n; \m_{\y|s_n}, \C_{\y|s_n}),
\end{eqnarray}
for the distorted speech $\y_n$, the likelihood in (\ref{eqn:hmm_totalsplit}) can be adapted according to
\begin{flalign*}
	&\hspace{3.2mm}p(\y_n|q_n) = \int p(\x_n|q_n) p(\y_n|\x_n) d\x_n&
\end{flalign*}
\vspace{-5mm}
\begin{eqnarray}
	\esp&=&\esp \int p(\x_n|q_n) \frac{p(\x_n,\y_n)}{p(\x_n)} d\x_n \nonumber \\
	\esp&=&\esp\int p(\x_n|q_n) \frac{\sum_{s_n} p(\x_n|\y_n,s_n)p(\y_n|s_n)p(s_n)}{\sum_{s_n} \int p(\x_n|\y_n,s_n)p(\y_n|s_n)p(s_n) d\y_n} d\x_n. \nonumber \\[-2mm] \label{eqn:splice_final}
\end{eqnarray}
Although analytically tractable, both the numerator and the denominator in \eq{eqn:splice_final} are typically approximated for the sake of runtime efficiency \cite{droppo_uncertainty_2002}.

\subsection{Joint Uncertainty Decoding}  \label{ssec:jud}

Model-based joint uncertainty decoding \cite{liao_uncertainty_2007} assumes an affine observation model in the cepstral domain
\begin{eqnarray}
	\y_n = \A_{k_n} \x_n + \b_n
\end{eqnarray}
with the deterministic matrix $\A_{k_n}$ and $p(\ww_n | k_n) = \NN(\ww_n ; \m_{\ww|k_n}, \C_{\ww|k_n})$ depending on the considered Gaussian component $k_n$
of the \gmm of the current \hmm state $q_n$:
\begin{eqnarray}
	p(\x_n | q_n) = \sum_{k_n} p(k_n) p(\x_n | k_n).
\end{eqnarray}
The Bayesian network is depicted in Figure~\ref{fig:vertall}(c) implying the following compensation rule:
\begin{eqnarray} \label{eqn:jud}
	p(\y_n|k_n) = \int p(\x_n | k_n) p(\y_n | \x_n, k_n) d\x_n,
\end{eqnarray}
which can be analytically derived analogously to (\ref{eqn:arrowood}).
In practice, the compensation parameters $\A_{k_n}, \m_{\b|k_n}, \C_{\b|k_n}$ are not estimated for each Gaussian component $k_n$
but for each regression class comprising a set of Gaussian components \cite{liao_uncertainty_2007}.

\subsection{REMOS} \label{ssec:remos}

\begin{figure}[!t]
	\centering
	\psfrag{e1}[l][l]{(a)}
	\psfrag{e2}[l][l]{(b)}
	\psfrag{A}[c][c]{$...$}
	\psfrag{E}[c][c]{$...$}
	\psfrag{q1}[l][l]{$q_{n\!-\!L}$}
	\psfrag{q2}[l][l]{$q_{n\!-\!1}$}
	\psfrag{q3}[l][l]{$q_n$}
	\psfrag{x1}[l][l]{$\x_{n\!-\!L}$}
	\psfrag{x2}[l][l]{$\x_{n\!-\!1}$}
	\psfrag{x3}[l][l]{$\x_n$}
	\psfrag{x4}[l][l]{$\overline{\x}_n$}
	\psfrag{z1}[l][l]{$\y_{n\!-\!L}$}
	\psfrag{z2}[l][l]{$\y_{n\!-\!1}$}
	\psfrag{z3}[l][l]{$\y_n$}
	\psfrag{h1}[l][l]{$\b_{n\!-\!L}$}
	\psfrag{h2}[l][l]{$\b_{n\!-\!1}$}
	\psfrag{h3}[l][l]{$\b_n$}%
	\psfrag{B}[l][l]{}
	\psfrag{C}[l][l]{}
	\includegraphics[width=0.8\columnwidth]{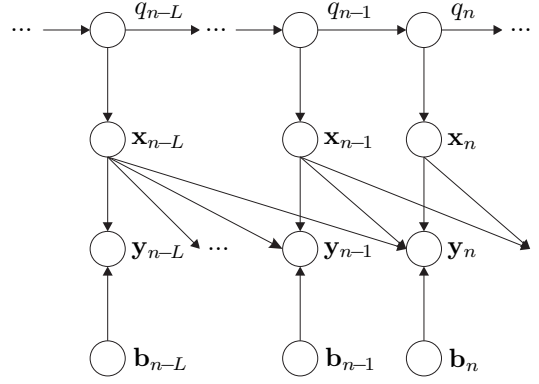}
	\caption{Bayesian network representation of the REMOS concept (Subsection~\mbox{\ref{sec:examples}-\ref{ssec:remos}}). The figure is based on \cite{maas_formulation_2013}.}
	\label{fig:remos}
\end{figure}

As many other techniques, %
the \remos concept \cite{sehr_reverberation_2010, maas_formulation_2013} assumes the environmental distortion to be additive in the melspectral domain.
However, REMOS also considers the influence of the $L$ previous clean speech feature vectors $\x_{n-L:n-1}$ %
in order to model the dispersive effect of reverberation and to relax the conditional independence assumption of conventional \hmms. The observation model reads in the \logmel domain:
\begin{eqnarray}
	\y_n \esp&=&\esp \log\bigg(\exp(\c_n) + \exp(\h_n + \x_n) \nonumber \\
	\esp&&\esp + \exp(\a_n) \; \sum_{l=1}^L{\exp({\boldsymbol\mu}_l + \x_{n-l})}\bigg), \label{eqn:remos}
\end{eqnarray}
where the normally distributed random variables $\c_n, \h_n, \a_n$ model the additive noise components,
the early part of the room impulse response (RIR), and the weighting of the late part of the RIR, respectively,
and the parameters ${\boldsymbol\mu}_{1:L}$ represent a deterministic description of the late part of the RIR.
The Bayesian network is depicted in Figure~\ref{fig:remos} with $\b_n = [\c_n, \a_n, \h_n]$.
In contrast to most of the other compensation rules discussed in this article,
the REMOS concept necessitates a modification of the Viterbi decoder due to the introduced cross-connections in Figure~\ref{fig:remos}.
In order to arrive at a computationally feasible decoder, the marginalization over the previous clean speech components $\x_{n-L:n-1}$
is circumvented by employing estimates $\widehat{\x}_{n-L:n-1}(q_{n-1})$ that depend on the best partial path, i.e., on the previous \hmm state $q_{n-1}$.
The resulting analytically intractable integral is then approximated by the maximum of its integrand:
\begin{flalign*}
	&\hspace{3.2mm}p(\y_n | q_n, \widehat{\x}_{n-L:n-1}(q_{n-1})) =&
\end{flalign*}
\vspace{-7mm}
\begin{eqnarray}
	&=& \int p(\y_{n}|\x_n,\widehat{\x}_{n-L:n-1}(q_{n-1})) p(\x_{n}|q_n) d\x_n \nonumber \\
	&\geq& \max_{\x_n}  \; p(\y_{n}|\x_n,\widehat{\x}_{n-L:n-1}(q_{n-1})) p(\x_{n}|q_n). \label{eqn:remosMAX}
\end{eqnarray}
The determination of a global solution to \eq{eqn:remosMAX} represents the core of the \remos concept.
The estimates $\widehat{\x}_{n-L:n-1}(q_{n-1})$ in turn are the solutions to \eq{eqn:remosMAX} at previous time steps.
We refer to \cite{maas_formulation_2013} for a detailed derivation of the corresponding decoding routine.

\subsection{Ion and Haeb-Umbach}  \label{ssec:ion} %

\begin{figure}[!t]
	\centering
	\psfrag{A}[c][c]{$...$}
	\psfrag{B}[l][l]{(a)}
	\psfrag{C}[l][l]{(b)}
	\psfrag{q2}[l][l]{$q_{n\!-\!1}$}
	\psfrag{q3}[l][l]{$q_n$}
	\psfrag{x2}[l][l]{$\x_{n\!-\!1}$}
	\psfrag{x3}[l][l]{$\x_n$}
	\psfrag{z2}[l][l]{$\y_{n\!-\!1}$}
	\psfrag{z3}[l][l]{$\y_n$}
	\includegraphics[width=\columnwidth]{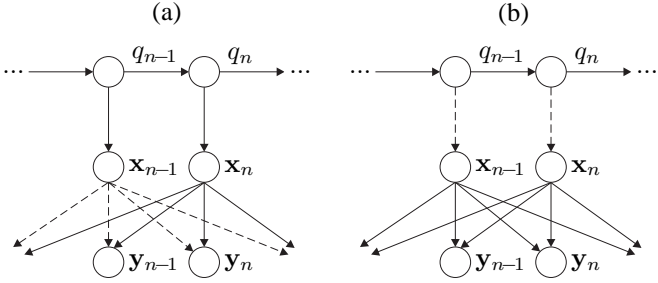}
	\caption{Bayesian network representation of the decoding rule of \cite{ion_novel_2008}, where the dashed links are disregarded in the different steps of the derivation (Subsection~\mbox{\ref{sec:examples}-\ref{ssec:ion}}).}
	\label{fig:ion}
\end{figure}

Similar to REMOS, the generic uncertainty decoding approach proposed by \cite{ion_novel_2008} considers cross-connections in the Bayesian network in order to relax the conditional independence assumption of \hmms.
The concept in \cite{ion_novel_2008} is an example of uncertainty decoding, where the compensation rule can be defined by a modified Bayesian network structure -- given in Figure~\ref{fig:ion}(a) --
without fixing a particular functional form of the involved pdfs via an analytical observation model.
In order to derive the compensation rule, %
we start by introducing the sequence $\x_{1:N}$ of latent clean speech vectors in each summand of (\ref{eqn:hmm}):
\begin{flalign*}
	&\hspace{3.2mm}p(\y_{1:N},q_{1:N}) = \int p(\y_{1:N},\x_{1:N},q_{1:N}) d\x_{1:N}&
\end{flalign*}
\vspace{-6mm}
\begin{eqnarray}
	\esp&=&\esp \int p(\y_{1:N}|\x_{1:N}) \; \prod_{n=1}^N p(\x_n|q_n) p(q_n|q_{n-1}) d\x_{1:N} \nonumber \\
	\esp&\sim&\esp \int \frac{p(\x_{1:N}|\y_{1:N})}{p(\x_{1:N})} \prod_{n=1}^N p(\x_n|q_n) \; p(q_n|q_{n-1}) d\x_{1:N}, \label{eqn:ion1}
\end{eqnarray}
where we exploited the conditional independence properties defined by Figure~\ref{fig:ion}(a) (respecting the dashed links)
and dropped $p(\y_{1:N})$ in the last line of (\ref{eqn:ion1}) as it represents a constant factor with respect to a varying state sequence $q_{1:N}$.
The pdf in the numerator of (\ref{eqn:ion1}) is next turned into
\begin{eqnarray}
	p(\x_{1:N}|\y_{1:N}) \esp&=&\esp p(\x_1|\y_{1:N}) \prod_{n=2}^N p(\x_n|\y_{1:N}, \x_{1:n-1}) \label{eqn:ion1num1} \nonumber \\
	\esp&\approx&\esp \prod_{n=1}^N p(\x_n|\y_{1:N}), \label{eqn:ion1num2}
\end{eqnarray}
where the conditional dependence (due to the head-to-head relation) of $\x_n$ and $\x_{1:n-1}$ is neglected.
This corresponds to omitting the respective dashed links in Figure~\ref{fig:ion}(a) for each factor in \eq{eqn:ion1num1} separately.
The denominator in (\ref{eqn:ion1}) can also be further decomposed if the dashed links in Figure~\ref{fig:ion}(b), i.e., the head-to-tail relations in $q_n$, are  disregarded:
\begin{equation}
	p(\x_{1:N}) \approx \prod_{n=1}^N  p(\x_n).\label{eqn:ion1num22}
\end{equation}
With \eq{eqn:ion1num2} and \eq{eqn:ion1num22}, the update rule (\ref{eqn:ion1}) is finally turned into the following simplified form:
\begin{equation} \label{eqn:ion2}
	p(\y_{1:N},q_{1:N}) \sim \prod_{n=1}^N  \int \frac{p(\x_n|\y_{1:N})}{p(\x_n)} \; p(\x_n|q_n) d\x_n \; p(q_n|q_{n-1})
\end{equation}
that is given in \cite{ion_novel_2008}. Due to the approximations in Figure~\ref{fig:ion}(a) and (b), the compensation rule defined by (\ref{eqn:ion2}) exhibits the same decoupling as (\ref{eqn:hmm_totalsplit})
and can thus be carried out without modifying the underlying decoder. %
In practice, $p(\x_n)$ may, e.g., be modeled as a separate Gaussian density and $p(\x_n | \y_{1:N})$ as a separate Markov process \cite{haeb_2011}.

\subsection{Feature Vector Imputation} \label{ssec:imputation}

We next turn to missing feature techniques, which can be used to model feature distortion due to a front-end enhancement process \cite{raj_missing-feature_2005}, noise \cite{cooke_missing_1997} or reverberation \cite{Palomaki04}.
A major subcategory of missing feature approaches is called {\em feature vector imputation} \cite{cooke_robust_2001, raj_missing-feature_2005, kolossa_robust_2011}
where each feature vector component $y_n^{(d)}$ is either classified as reliable ($d \in \mathcal{R}_n$)
or unreliable ($d \in \mathcal{U}_n$), where $\mathcal{R}_n$ and $\mathcal{U}_n$ denote the set of reliable and unreliable components of the $n$-th feature vector, respectively \cite{haeb_2011}.
While unreliable components are withdrawn and replaced by an estimate $\widehat{x}_n^{(d)}$ of the original observation $y^{(d)}_n$,
reliable components are directly \qm{plugged} into the pdf. The score calculation in (\ref{eqn:hmm_totalsplit}) therefore becomes
\begin{eqnarray}
	\p(\y_n|q_n) \esp&=&\esp \int p(\x_n|q_n) \frac{p(\x_n|\y_n)}{p(\x_n)} d\x_n \nonumber \\
	\esp&\approx&\esp \int p(\x_n|q_n) p(\x_n|\y_n) d\x_n \label{eqn:noDenom}
\end{eqnarray}
with 
\begin{eqnarray}
	p(\x_n|\y_n) = \prod_{d=1}^D p(x_n^{(d)}|y_n^{(d)})
\end{eqnarray}
and \cite{haeb_2011}
\begin{eqnarray} \label{eqn:imputation}
	p(x_n^{(d)}|y_n^{(d)}) =
	\left\{
	\begin{array}{l l}
		\delta(x_n^{(d)} - y_n^{(d)}) & d \in \mathcal{R} \\
		\delta(x_n^{(d)} - \widehat{x}_n^{(d)}) & d \in \mathcal{U}
	\end{array}
	\right.
\end{eqnarray}
with the general Bayesian network in Figure~\ref{fig:vertdet}(a).

\subsection{Marginalization} \label{ssec:marginalization}

The second major subcategory of missing feature techniques is called {\em marginalization} \cite{cooke_robust_2001, raj_missing-feature_2005, kolossa_robust_2011}, where unreliable components
are \qm{replaced} by marginalizing over a clean-speech distribution $p(x_n^{(d)})$ that is usually not derived from the \hmm but separately modeled.
The posterior likelihood in (\ref{eqn:imputation}) thus becomes \cite{haeb_2011}
\begin{eqnarray}
	p(x_n^{(d)}|y_n^{(d)}) =
	\left\{
	\begin{array}{l l}
		\delta(x_n^{(d)} - y_n^{(d)}) & d \in \mathcal{R} \\
		p(x_n^{(d)}) & d \in \mathcal{U}
	\end{array}
	\right.
\end{eqnarray}
with the general Bayesian network in Figure~\ref{fig:vertdet}(a).

\subsection{Modified Imputation} \label{ssec:mod_imp}

The approach presented in \cite{kolossa_separation_2005} again implicitly assumes the basic observation model $\y_n = \x_n + \b_n$,
where $\b_n \sim \NN(\0, \C_{\b_n})$ denotes the uncertainty of the enhancement algorithm.
The corresponding Bayesian network is depicted in Figure~\ref{fig:vertall}(a) implying that
the observation likelihood in (\ref{eqn:noDenom}) becomes
\begin{eqnarray}
	\p(\y_n|q_n) \esp&\approx&\esp \int p(\x_n|q_n) p(\x_n|\y_n) d\x_n \nonumber \\
	\esp&\approx&\esp \int p(\x_n|q_n) \delta(\x_n-\widehat{\x}_n) d\x_n, \label{eqn:mi}
\end{eqnarray}
where $\widehat{\x}_n = \argmax_{\x_n} p(\x_n|q_n) p(\x_n|\y_n)$.
Note that if $\widehat{\x}_n$ were determined to be the argmax of $p(\x_n|\y_n)$, it could be considered a \map \cite{bishop_pattern_2006} estimate.

\subsection{Significance Decoding} \label{ssec:sig_dec}

The concept of significance decoding \cite{abdelaziz_decoding_2012} arises directly from (\ref{eqn:mi}) by approximating the integral by its maximum integrand:
\begin{eqnarray}
	\p(\y_n|q_n) \esp&\approx&\esp \int p(\x_n|q_n) p(\x_n|\y_n) d\x_n \nonumber \\
	\esp&\geq&\esp \max p(\x_n|q_n) p(\x_n|\y_n). \label{eqn:si}
\end{eqnarray}
Also this concept considers the observation uncertainty to be provided by an acoustic front-end.

\begin{figure}[!t]
	\centering
	\psfrag{A}[c][c]{$...$}
	\psfrag{e2}[l][l]{(a)}
	\psfrag{e4}[l][l]{(b)}
	\psfrag{e5}[l][l]{(c)}
	\psfrag{a1}[l][l]{$q_{n\!-\!1}$}
	\psfrag{a2}[l][l]{$q_n$}
	\psfrag{x2}[l][l]{$\x_n$}
	\psfrag{x1}[l][l]{$\x_{n\!-\!1}$}
	\psfrag{c1}[l][l]{$\y_{n\!-\!1}$}
	\psfrag{c2}[l][l]{$\y_n$}
	\psfrag{b2}[l][l]{$\b_n$}
	\psfrag{d2}[l][l]{$s_n$}
	\psfrag{a3}[l][l]{$\q_n$}
	\psfrag{q2}[l][l]{$q_{n\!-\!1}$}
	\psfrag{q3}[l][l]{$q_n$}
	\psfrag{k2}[l][l]{$k_n$}
	\psfrag{y2}[l][l]{$\y_{n\!-\!1}$}
	\psfrag{y3}[l][l]{$\y_n$}
	\psfrag{b1}[l][l]{$\th$}
	\includegraphics[width=\columnwidth]{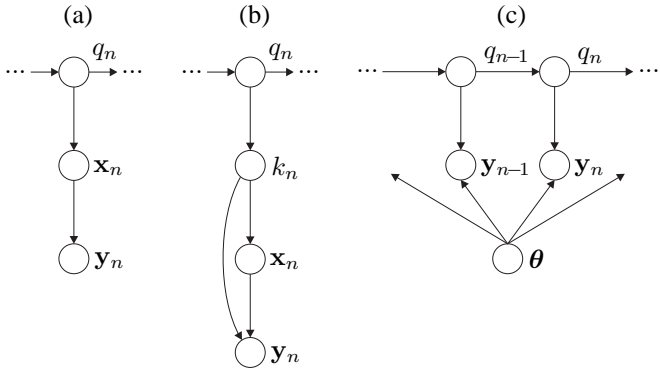}
	\caption{Bayesian network representation of (a) different model compensation techniques, (b) CMLLR (Subsection~\mbox{\ref{sec:examples}-\ref{ssec:cmllr}}) and MLLR (Subsection~\mbox{\ref{sec:examples}-\ref{ssec:mllr}}), and (c) MAP adaptation (Subsection~\mbox{\ref{sec:examples}-\ref{ssec:map}}).
	Detailed descriptions are given in the text.}
	\label{fig:vertdet}
\end{figure}

\subsection{Parallel Model Combination} \label{ssec:pmc} %

We next investigate several acoustic model adaptation techniques starting with the fundamental framework of \pmc \cite{gales_model-based_1995}.
The observation model of the \pmc concept
is based on the log-sum distortion model %
and reads in the static \logmel domain:
\begin{eqnarray} \label{eqn:pmc_model}
	\y_n = \log(\alpha \exp(\x_n) + \exp(\b_n)),
\end{eqnarray}
where the deterministic parameter $\alpha$ accounts for level differences between the clean speech $\x_n$ and the distortion $\b_n$.
Under the assumption of stationary distortions, i.e.,
\begin{eqnarray}
 	p(\b_n) = {\rm const.},
\end{eqnarray}
the underlying Bayesian network corresponds to Figure~\ref{fig:vertall}(a).
This explains the name of \pmc as \eq{eqn:pmc_model} combines two independent parallel models: the clean-speech \hmm and the distortion model $p(\ww_n)$.
Since the resulting adapted pdf
\begin{eqnarray}
	p(\y_n|q_n) = \int p(\x_n|q_n) p(\y_n|\x_n, \b_n) p(\b_n) d(\x_n,\b_n) \label{eqn:pmc}
\end{eqnarray}
cannot be derived in an analytical closed form, a variety of approximations to the true pdf $p(\y_n | q_n)$ have been investigated \cite{gales_model-based_1995}.
For nonstationary distortions, \cite{gales_model-based_1995} proposes to employ a separate \hmm for the distortion $\ww_n$
leading to the Bayesian network representation of Figure~\ref{fig:vertall}(d).
Marginalizing over the distortion state sequence $\q_{1:N}$ as in (\ref{eqn:hmm_totalsplit}) reveals the acoustic score to become %
\begin{flalign*}
	&\hspace{3.2mm}p(\y_{1:N} | \W) = \sum_{\substack{q_{1:N}\\\q_{1:N}}} p(\y_{1:N},q_{1:N},\q_{1:N})&
\end{flalign*}
\vspace{-4mm}
\begin{eqnarray}
	\esp&=&\esp \sum_{\substack{q_{1:N}\\\q_{1:N}}} \bigg\{\prod_{n=1}^N p(\y_n|q_n,\q_n) \; p(q_n|q_{n-1}) \; p(\q_n|\q_{n-1})\bigg\},
\end{eqnarray}
where
\begin{eqnarray}
	p(\y_n|q_n, \q_n) \esp&=&\esp \int p(\x_n | q_n) p(\y_n | \x_n, \b_n) p(\b_n | \q_n) d(\x_n,\b_n). \nonumber \\[-2mm] \label{eqn:noiseHMM}
\end{eqnarray}
The overall acoustic score can be approximated by a 3D Viterbi decoder,
which can in turn be mapped onto a conventional 2D Viterbi decoder \cite{gales_model-based_1995}.

\subsection{Vector Taylor Series} \label{ssec:vts} %

The concept of \vts is frequently employed in practice yielding promising results \cite{li_overview_2014}.
Its fundamental idea is to linearize a nonlinear distortion model by a Taylor series \cite{moreno_speech_1996, acero_hmm_2000, li_unified_2009}.
The standard \vts approach \cite{moreno_speech_1996} is based on the log-sum observation model:
\begin{eqnarray} \label{eqn:vts}
	\y_n = \log(\exp(\h_n + \x_n) + \exp(\c_n)),
\end{eqnarray}
where $p(\h_n) = \NN(\h_n ; \m_\h, \C_\h)$ captures short convolutive distortion and %
$p(\c_n) = \NN(\c_n ; \m_\c, \C_\c)$ models additive noise components. The Bayesian network is represented by Figure~\ref{fig:vertall}(a) with $\b_n = [\h_n, \c_n]$.
Note that in contrast to uncertainty decoding, $p(\b_n)$ is constant over time.
As the adapted pdf is again of the form of (\ref{eqn:pmc}) and thus analytically intractable,
it is assumed that (\ref{eqn:vts}) can, firstly, be applied to each Gaussian component $p(\x_n | k_n)$ of the \gmm
\begin{eqnarray}
	p(\x_n | q_n) = \sum_{k_n} p(k_n) p(\x_n | k_n)
\end{eqnarray}
individually and, secondly, be approximated by a Taylor series around $[\m_{\x|k_n}, \m_\h, \m_\c]$,
where $\m_{\x|k_n}$ denotes the mean of the component $p(\x_n | k_n)$.
There are various extensions to the \vts concept that are neglected here. For a more comprehensive review of \vts, we refer to \cite{li_overview_2014}.

\subsection{CMLLR} \label{ssec:cmllr}

\cmllr \cite{digalakis_speaker_1995, gales_maximum_1997} can be seen as the deterministic counterpart of joint uncertainty decoding (Subsection~\mbox{\ref{sec:examples}-\ref{ssec:jud}})
with the observation model
\begin{eqnarray}
	\y_n = \A_{k_n} \x_n + \b_{k_n} \label{eqn:cmllrObsModel}
\end{eqnarray}
and deterministic parameters $\A_{k_n}, \b_{k_n}$.
The adaptation rule of $p(\y_n|k_n)$ has the same form as (\ref{eqn:jud}) with
\begin{eqnarray} \label{eqn:cmllrDirac}
	p(\y_n | \x_n, k_n) = \delta(\y_n - \A_{k_n} \x_n - \b_{k_n}).
\end{eqnarray}
The Bayesian network corresponds to Figure~\ref{fig:vertdet}(b),
where the use of regression classes is again reflected by the dependency of the observation model parameters on the Gaussian component $k_n$ (cf. Subsection~\mbox{\ref{sec:examples}-\ref{ssec:jud}}).
The affine observation model in \eq{eqn:cmllrObsModel} is equivalent to transforming the mean vector $\m_{\x|k_n}$ and covariance matrix $\C_{\x|k_n}$ of each Gaussian component of $p(\x_n|k_n)$:
\begin{eqnarray}
	\m_{\y|k_n} \esp&=&\esp \A_{k_n} \m_{\x|k_n} + \ww_{k_n}, \label{eqn:cmllrTrafo1} \\
	\C_{\y|k_n} \esp&=&\esp \A_{k_n} \C_{\x|k_n} \A_{k_n}^T. \label{eqn:cmllrTrafo2}
\end{eqnarray}
\cmllr represents a very popular adaptation technique in practice
due to its promising results and versatile fields of application,
such as speaker adaptation \cite{gales_maximum_1997}, adaptive training \cite{kai_yu_discriminative_2006} as well as noise \cite{vincent_secondchimespeech_2013} and reverberation-robust \cite{kinoshita_reverb_2013} \asr.

\subsection{MLLR} \label{ssec:mllr}

The \mllr concept \cite{leggetter_maximum_1995} can be considered as a generalization of \cmllr as it allows for a separate transform matrix $\B_{k_n}$ in \eq{eqn:cmllrTrafo2}:
\begin{eqnarray}
	\m_{\y|k_n} \esp&=&\esp \A_{k_n} \m_{\x|k_n} + \ww_{k_n}, \label{eqn:mllrTrafo1} \\
	\C_{\y|k_n} \esp&=&\esp \B_{k_n} \C_{\x|k_n} \B_{k_n}^T. \label{eqn:mllrTrafo2}
\end{eqnarray}
In practice, however, \mllr is frequently applied to the mean vectors only \cite{anastasakos_compact_1996, young_htk_2002, delcroix_speech_2011, xiong_robust_2014}
while neglecting the adaptation of the covariance matrix:
\begin{eqnarray}
	\C_{\y|k_n} = \C_{\x|k_n}.
\end{eqnarray}
This principle is also known from other approaches
that are applicable to both means and variances but are often only carried out on the former (e.g., for the sake of robustness) \cite{gales_model-based_1995, hirsch_hmm_2001}.

If applied to the mean vectors only, \mllr can in turn be considered as a simplified version of \cmllr,
where the observation model \eq{eqn:cmllrObsModel} and the Bayesian network in Figure~\ref{fig:vertdet}(b) is assumed
while the compensation of the variances %
is omitted.

The Bayesian network representation in Figure~\ref{fig:vertdet}(b) also underlies the general \mllr adaptation rule \eq{eqn:mllrTrafo1} and \eq{eqn:mllrTrafo2}. In this case, however, it seems impossible to identify a corresponding analytic observation model representation.

\subsection{MAP Adaptation} \label{ssec:map}

We next describe the MAP adaptation applied to any parameters $\th$ of the pdfs of an \hmm. %
In MAP, these parameters are considered as Bayesian parameters, i.e., random variables that are drawn once for all times
as depicted in Figure~\ref{fig:vertdet}(c) \cite{bishop_pattern_2006}.
As a direct consequence, any two observation vectors $\y_i, \y_j$ are conditionally dependent given the state sequence.
The predictive pdf in (\ref{eqn:hmm}) therefore explicitly depends on the adaptation data that we denote as $\y_{M:0}$, $M < 0$,
and becomes
\begin{eqnarray}
	p(\y_{1:N},q_{1:N} | \y_{M:0}) \esp&=&\esp \int p(\y_{1:N},q_{1:N}, \th | \y_{M:0}) d\th \nonumber\\
	\esp&=&\esp \int p(\y_{1:N},q_{1:N} | \th, \y_{M:0}) p(\th | \y_{M:0}) d\th \nonumber\\
	\esp&\approx&\esp p(\y_{1:N},q_{1:N} | \th_{\rm MAP}, \y_{M:0}), \label{eqn:map1}
\end{eqnarray}
where the posterior $p(\th | \y_{M:0})$ is approximated as Dirac distribution $\delta(\th - \th_{\rm MAP})$ at the mode $\th_{\rm MAP}$:
\begin{equation} \label{eqn:map2}
	\th_{\rm MAP} = \argmax_{\th} p(\th | \y_{M:0}) = \argmax_{\th} p(\y_{M:0} | \th) p(\th).
\end{equation}
An iterative (local) solution to (\ref{eqn:map2}) is obtained by the \exmax algorithm.
Note that due to the MAP approximation of the posterior $p(\th|\y_{M:0})$, the conditional independence assumption is again fulfilled
such that a conventional decoder can be employed.

\begin{figure}[!t]
	\centering
	\psfrag{A}[c][c]{$...$}
	\psfrag{B}[l][l]{(a)}
	\psfrag{C}[l][l]{(b)}
	\psfrag{q2}[l][l]{$q_{n\!-\!1}$}
	\psfrag{q3}[l][l]{$q_n$}
	\psfrag{y2}[l][l]{$\y_{n\!-\!1}$}
	\psfrag{y3}[l][l]{$\y_n$}
	\psfrag{x2}[l][l]{$\x_{n\!-\!1}$}
	\psfrag{x3}[l][l]{$\x_n$}
	\psfrag{b2}[l][l]{$\b_{n\!-\!1}$}
	\psfrag{b3}[l][l]{$\b_n$}
	\psfrag{b1}[l][l]{$\b$}
	\includegraphics[width=\columnwidth]{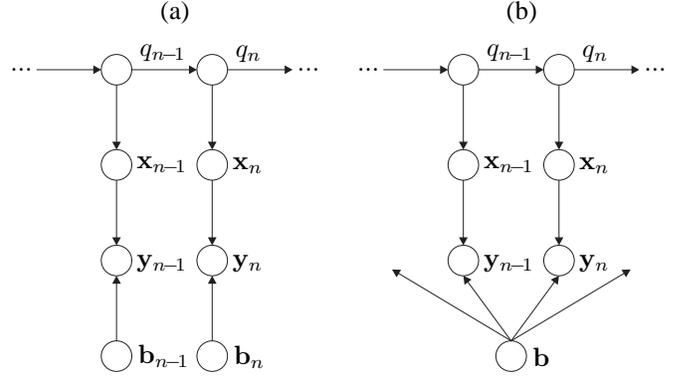}
	\caption{Bayesian networks representing (a) typical uncertainty decoding and model adaptation with probabilistic parameter $\b_n$
	and (b) Bayesian model adaptation.}
	\label{fig:Bayes_vs_UncDec}
\end{figure}

\subsection{Bayesian MLLR} \label{ssec:bayes_mllr}

As mentioned before, uncertainty decoding techniques allow for a time-varying pdf $p(\ww_n)$,
while model adaptation approaches, such as in Subsections~\mbox{\ref{sec:examples}-\ref{ssec:pmc}}, \mbox{\ref{sec:examples}-\ref{ssec:vts}}, and \mbox{\ref{sec:examples}-\ref{ssec:rev_vts}}, mostly set $p(\ww_n)$ to be constant over time.
In both cases, however, the \qm{randomized} model parameter $\ww_n$ is assumed to be redrawn in each time step $n$ as in Figure~\ref{fig:Bayes_vs_UncDec}(a).
In contrast, {\em Bayesian estimation} -- as mentioned before -- usually refers to inference problems,
where the random model parameters are drawn once for all times \cite{bishop_pattern_2006} as in Figure~\ref{fig:Bayes_vs_UncDec}(b).

Another example of Bayesian model adaptation, besides MAP, is Bayesian \mllr \cite{chien_linear_2003} applied to the mean vector $\m_{\x|q_n}$ of each pdf $p(\x_n|q_n)$:
\begin{eqnarray} \label{eqn:bayesianMLLR}
	\m_{\y|q_n} = \A \m_{\x|q_n} + \c
\end{eqnarray}
with $\ww = [\A, \c]$ being usually drawn from a Gaussian distribution \cite{chien_linear_2003}.
Here, we do not consider different regression classes and assume $p(\x_n | q_n)$ to be a single Gaussian pdf
since, in the case of a \gmm, the linear mismatch function (\ref{eqn:bayesianMLLR}) can be applied to each Gaussian component separately.
The likelihood score in (\ref{eqn:hmm}) thus becomes\footnote{In contrast to (\ref{eqn:map1}),
we do not explicitly mention the dependency on the adaptation data $\y_{M:0}$ for notational convenience.}:
\begin{eqnarray}
	\sum_{q_{1:N}} p(\y_{1:N},q_{1:N}) \esp&=&\esp \sum_{q_{1:N}} \int p(\y_{1:N},q_{1:N}, \b) d\b \nonumber \\
	\esp&=&\esp \sum_{q_{1:N}} \int p(\y_{1:N},q_{1:N} | \b) p(\b) d\b. \label{eqn:BayesianInt}
\end{eqnarray}
This score can, e.g., be approximated by a frame-synchronous Viterbi search \cite{jiang_robust_1999}.
Another approach is to apply the Bayesian integral in a frame-wise manner and use a standard decoder \cite{jen-tzung_chien_linear_2003}.
In this case, the integral in (\ref{eqn:BayesianInt}) becomes
\begin{flalign*}
	&\int p(\y_{1:N},q_{1:N} | \ww) p(\ww) d\ww =&
\end{flalign*}
\vspace{-6mm}
\begin{eqnarray}
	&=&\esp \int \prod_{n=1}^N  p(\y_n| q_n, \ww) \; p(q_n|q_{n-1})\; p(\ww) d\ww \nonumber \\
	&\approx&\esp \prod_{n=1}^N \int p(\y_n| q_n, \ww) \; p(q_n|q_{n-1}) \; p(\ww) d\ww, \label{eqn:bayesMLLR2}
\end{eqnarray}
where the original assumption of $\w$ being {\em identical} for all time steps $n$ was relaxed
to the case of $\w$ being {\em identically distributed} for all times steps $n$.
The approximation in \eq{eqn:bayesMLLR2} is representable
by the conversion of the Bayesian network in Figure~\ref{fig:Bayes_vs_UncDec}(b) to the one in (a) with constant \pdf $p(\ww_n) = p(\ww)$ for all $n$. %

\begin{figure}[!t]
	\centering
	\psfrag{A}[c][c]{$...$}
	\psfrag{B}[l][l]{(a)}
	\psfrag{C}[l][l]{(b)}
	\psfrag{q2}[l][l]{$q_{n\!-\!1}$}
	\psfrag{q3}[l][l]{$q_n$}
	\psfrag{y2}[l][l]{$\y_{n\!-\!1}$}
	\psfrag{y3}[l][l]{$\y_n$}
	\psfrag{x2}[l][l]{$\x_{n\!-\!1}$}
	\psfrag{x3}[l][l]{$\x_n$}
	\psfrag{z2}[l][l]{$\overline{\x}_{n\!-\!1}$}
	\psfrag{z3}[l][l]{$\overline{\x}_n$}
	\psfrag{b2}[l][l]{$\b_{n\!-\!1}$}
	\psfrag{b3}[l][l]{$\b_n$}
	\psfrag{b1}[l][l]{$\m$}
	\includegraphics[width=\columnwidth]{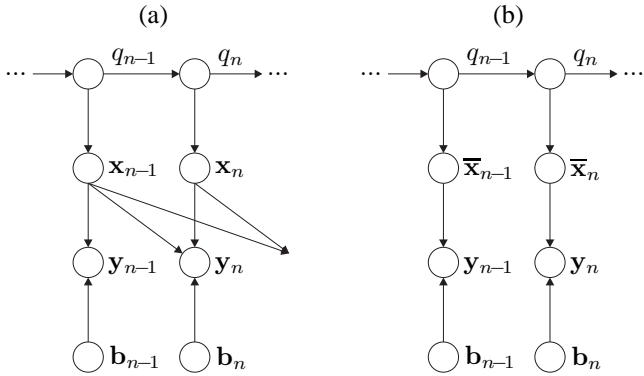}
	\caption{Bayesian network representation of reverberant VTS (Subsection~\mbox{\ref{sec:examples}-\ref{ssec:rev_vts}}) (a) before and (b) after approximation via an extended observation vector.
	The figure is based on \cite{wang_improving_2011}.}
	\label{fig:rev_adapt}
\end{figure}

\subsection{Reverberant VTS} \label{ssec:rev_vts}

Reverberant VTS \cite{wang_improving_2011} is an extension of conventional VTS (Subsection~\mbox{\ref{sec:examples}-\ref{ssec:vts}}) to capture the dispersive effect of reverberation. Its observation model reads for static features in the logmelspec domain:
\begin{eqnarray} \label{eqn:revVTS}
	\y_n = \log\bigg(\sum_{l=0}^L \exp(\x_{n-l} + \m_l) + \exp(\b_n)\bigg)
\end{eqnarray}
with $\b_n$ being an additive noise component modeled as normally distributed random variable and $\m_{0:L}$ being a deterministic description of the reverberant distortion.
For the sake of tractability, the observation model is approximated in a similar manner as in the VTS approach.
This concept can be seen as an alternative to REMOS (Subsection~\mbox{\ref{sec:examples}-\ref{ssec:remos}}): While REMOS tailors the Viterbi decoder to the modified Bayesian network,
reverberant VTS avoids the computationally expensive marginalization over all previous clean-speech vectors %
by averaging -- and thus smoothing -- the clean-speech statistics over all possible previous states and Gaussian components.
Thus, $\y_n$ is assumed to depend on the extended clean-speech vector $\overline{\x}_n$ = $[\x_{n-L},...,\x_n]$, cf.\ Figure~\ref{fig:rev_adapt}(a) vs. (b).

\subsection{Convolutive Model Adaptation} \label{ssec:hirsch_raut_sehr}

Besides the previously mentioned \remos, reverberant \vts, and reverberant \cmllr concepts,
there are three related approaches employing a convolutive observation model in order
to describe the dispersive effect of reverberation \cite{hirsch_new_2008, raut_model_2006, sehr_frame-wise_2011}.
All three approaches assume the following model in the \logmel domain:
\begin{eqnarray} \label{eqn:hirsch_raut_sehr}
	\y_n = \log\bigg(\sum_{l=0}^L \exp(\x_{n-l} + \m_l)\bigg),
\end{eqnarray}
where $\m_{0:L}$ denotes a deterministic description of the reverberant distortion that is differently determined by the three approaches.
The observation model (\ref{eqn:hirsch_raut_sehr}) can be represented by the Bayesian network in Figure~\ref{fig:remos} without the random component $\b_n$.
Both \cite{hirsch_new_2008} and \cite{sehr_frame-wise_2011} use the \qm{log-add approximation} \cite{gales_model-based_1995} to derive $p(\y_n | q_n)$, i.e.,
\begin{eqnarray}
	\m_{\y|k_n} \esp&=&\esp \log\bigg(\exp(\m_{\x|k_n} + \m_0) \nonumber \\
		&&\esp + \sum_{l=1}^L \exp(\overline{\m}_{\x|q_{n-l}} + \m_l)\bigg),
\end{eqnarray}
where $\m_{\y|k_n}$ and $\m_{\x|k_n}$ denote the mean of the $k_n$-th Gaussian component of $p(\y_n|q_n)$ and $p(\x_n|q_n)$, respectively. %
The previous means $\overline{\m}_{\x|q_{n-l}}$, $l>0$,
are averaged over all means of the corresponding \gmm $p(\x_{n-l} | q_{n-l})$.
On the other hand, \cite{raut_model_2006} employs the \qm{log-normal approximation} \cite{gales_model-based_1995} to adapt $p(\y_n|q_n)$ according to (\ref{eqn:hirsch_raut_sehr}).
While \cite{hirsch_new_2008} and \cite{raut_model_2006} perform the adaptation once prior to recognition and then use a standard decoder,
the concept proposed in \cite{sehr_frame-wise_2011} performs an online adaptation based on the best partial path \cite{wang_improving_2011}.

It should be pointed out here that there is a variety of other approximations to the statistics of the log-sum of (mixtures of) Gaussian random variables
(as seen in Subsections~\ref{ssec:dvc}, \ref{ssec:remos}, \ref{ssec:pmc}, \ref{ssec:vts}, \ref{ssec:rev_vts} of Section~\ref{sec:examples}),
ranging from different PMC methods \cite{gales_model-based_1995} %
to maximum \cite{nadas_speech_1989}, piecewise linear \cite{maas_highly_2010},
and other analytical approximations \cite{schwartz_distribution_1982, beaulieu_estimating_1995, raut_2004, beaulieu_optimal_2004, hershey_2013}.

\begin{figure}[!t]
	\centering
	\psfrag{A}[c][c]{$...$}
	\psfrag{q2}[l][l]{$q_{n\!-\!1}$}
	\psfrag{q3}[l][l]{$q_n$}
	\psfrag{z2}[l][l]{$\x_{n\!-\!1}$}
	\psfrag{z3}[l][l]{$\x_n$}
	\psfrag{y2}[l][l]{$\y_{n\!-\!1}$}
	\psfrag{y3}[l][l]{$\y_n$}
	\includegraphics[width=0.5\columnwidth]{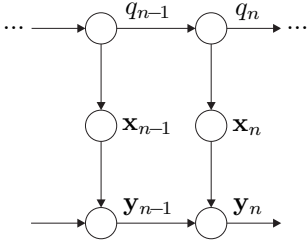}
	\caption{Bayesian network representation of \cite{takiguchi_acoustic_2006} (Subsection~\mbox{\ref{sec:examples}-\ref{ssec:takiguchi}}).}
	\label{fig:taki}
\end{figure}

\subsection{Takiguchi et al.} \label{ssec:takiguchi}

In contrast to the approaches of Subsections~\mbox{\ref{sec:examples}-\ref{ssec:rev_vts}} and \mbox{\ref{sec:examples}-\ref{ssec:hirsch_raut_sehr}}, the concept proposed in \cite{takiguchi_acoustic_2006}
assumes the reverberant observation vector $\y_{n-1}$ at time $n-1$ to be an approximation to the reverberation tail at time $n$ in the \logmel domain:
\begin{equation} \label{eqn:takiguchi}
	\y_n = \log\bigg(\exp(\h + \x_n) + \exp(\balpha + \y_{n-1})\bigg),
\end{equation}
where $\h$ and $\balpha$ are deterministic parameters modeling short convolutive distortion and the weighting of the reverberation tail, respectively.
Thus, each summand in (\ref{eqn:hmm}) becomes
\begin{eqnarray}
	p(\y_{1:N}, q_{1:N}) = \prod_{n=1}^N p(\y_n | q_n, \y_{n-1}) p(q_n | q_{n-1})
\end{eqnarray}
with the Bayesian network of Figure~\ref{fig:taki}.
It seems interesting to note that (\ref{eqn:takiguchi}) can be analytically evaluated as $\y_{n-1}$ is observed
and, thus, (\ref{eqn:takiguchi}) represents a nonlinear mapping $f(\cdot)$ of one random vector $\x_n$:
$\y_n = f(\x_n)$ with
\begin{eqnarray} \label{eqn:takiguchiPDF}
	p(\y_n | q_n, \y_{n-1}) = \frac{p(\x_n|q_n)}{\det(J_{\y_n}(f^{-1}(\y_n))},
\end{eqnarray}
where $\x_n = f^{-1}(\y_n)$ and $J_{\y_n}$ denotes the Jacobian w.r.t.\ $\y_n$.

\subsection{Conditional \hmms \cite{ming_modelling_1996} and Combined-Order \hmms \cite{maas_combined-order_2012}} \label{ssec:cond_comb_hmm}

We close this section by turning to two model-based approaches that cannot be classified as \qm{model adaptation}
as they postulate different \hmm topologies rather than adapting a conventional \hmm.
Both approaches aim at relaxing the conditional independence assumption of conventional \hmms
in order to improve the modeling of the inter-frame correlation. %

\begin{figure}[!t]
	\centering
	\psfrag{A}[c][c]{$...$}
	\psfrag{B}[l][l]{(a)}
	\psfrag{C}[l][l]{(b)}
	\psfrag{q2}[l][l]{$q_{n\!-\!1}$}
	\psfrag{q3}[l][l]{$q_n$}
	\psfrag{y2}[l][l]{$\y_{n\!-\!1}$}
	\psfrag{y3}[l][l]{$\y_n$}
	\includegraphics[width=\columnwidth]{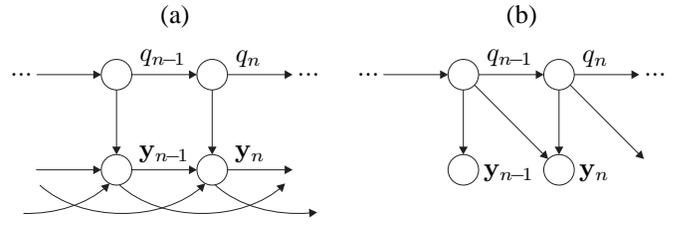}
	\caption{Bayesian network representation of (a) conditional and (b) combined-order \hmms (Subsection~\mbox{\ref{sec:examples}-\ref{ssec:cond_comb_hmm}}).
	Subfigure (a) is based on \cite{bishop_pattern_2006}.}
	\label{fig:HMM_CO_cond}
\end{figure}

The concept of conditional \hmms \cite{ming_modelling_1996} models the observation $\y_n$ as depending on the previous observations at time shifts $\boldsymbol\psi = (\psi_1, ..., \psi_P) \in \mathbb{N}^P$.
Each summand in (\ref{eqn:hmm}) therefore becomes
\begin{eqnarray}
	p(\y_{1:N}, q_{1:N}) \esp&=&\esp \prod_{n=1}^N p(\y_n| \y_{n-\psi_1}, ..., \y_{n-\psi_P}, q_n) \nonumber \\
	\esp&&\esp \cdot \; p(q_n | q_{n-1})
\end{eqnarray}
according to Figure~\ref{fig:HMM_CO_cond}(a). Such \hmms are also known as {\em autoregressive \hmms} \cite{bishop_pattern_2006}.

In contrast to conditional \hmms, combined-order \hmms \cite{maas_combined-order_2012} assume the current observation $\y_n$ to depend on the previous \hmm state $q_{n-1}$ in addition to state $q_n$:
\begin{equation}
	p(\y_{1:N}, q_{1:N}) = \prod_{n=1}^N p(\y_n| q_n, q_{n-1}) p(q_n | q_{n-1})
\end{equation}
according to Figure~\ref{fig:HMM_CO_cond}(b).

\section{Summary} \label{sec:conclusion}

In this article, we described the compensation rules of several acoustic model-based techniques employing Bayesian network representations.
Some of the presented Bayesian network descriptions are already given in the original papers
and others can be easily derived based on the original papers (cf.\ Subsections \mbox{\ref{sec:examples}-\ref{ssec:splice}}, \mbox{\ref{sec:examples}-\ref{ssec:jud}}, \mbox{\ref{sec:examples}-\ref{ssec:ion}}, and \mbox{\ref{sec:examples}-\ref{ssec:cond_comb_hmm}}).
In contrast, the link of the concepts of modified imputation (Subsection \mbox{\ref{sec:examples}-\ref{ssec:mod_imp}}), significance decoding (Subsection \mbox{\ref{sec:examples}-\ref{ssec:sig_dec}}),
\cmllr/MLLR (Subsections \mbox{\ref{sec:examples}-\ref{ssec:cmllr}} and \mbox{\ref{sec:examples}-\ref{ssec:mllr}}), MAP (Subsection \mbox{\ref{sec:examples}-\ref{ssec:map}}), Bayesian MLLR (Subsection \mbox{\ref{sec:examples}-\ref{ssec:bayes_mllr}}),
and Takiguchi et al. \cite{takiguchi_acoustic_2006} (Subsection \mbox{\ref{sec:examples}-\ref{ssec:takiguchi}})
to the Bayesian network framework via the formulations in \eq{eqn:mi}, \eq{eqn:si}, \eq{eqn:cmllrDirac}, \eq{eqn:map1}, \eq{eqn:bayesMLLR2}, and \eq{eqn:takiguchiPDF}, respectively,
are explicitly stated for the first time in this paper.

Clearly, the graphical model description neglects various crucial aspects related to the considered concepts:
Most importantly, neither the particular functional form of the joint pdf,
nor potential approximations to arrive at a tractable algorithm nor the provenance of (i.e., the estimation procedure for) the compensation parameters are reflected.
On the other hand, the Bayesian network description provides a convenient language to immediately and clearly
identify some major properties:
\begin{itemize}
	\item The cross-connections depicted in Figures~\ref{fig:remos}, \ref{fig:ion}, \ref{fig:rev_adapt}(a), \ref{fig:taki}, \ref{fig:HMM_CO_cond}
		show
		that the underlying concept aims at improving the modeling of the inter-frame correlation,
		e.g., to increase the robustness of the acoustic model against reverberation.
		If applied in a straightforward way, such cross-connections would entail a costly modification of the Viterbi decoder.
		In this paper, we summarized some important approximations that allow for a more efficient decoding of the extended Bayesian network,
		cf.\ Subsections~\mbox{\ref{sec:examples}-\ref{ssec:remos}}, \mbox{\ref{sec:examples}-\ref{ssec:ion}}, \mbox{\ref{sec:examples}-\ref{ssec:rev_vts}}, \mbox{\ref{sec:examples}-\ref{ssec:hirsch_raut_sehr}}.
		Some of these typically empirically motivated or just intuitive approximations, especially neglected statistical dependencies, become obvious from a Bayesian network, as shown in Figures~\ref{fig:ion} and \ref{fig:rev_adapt}.
	\item The approaches introducing instantaneous (here: purely vertical) extensions to the Bayesian network,
		as in Figures~\ref{fig:vertall}(a)-(c) and \ref{fig:vertdet}(c), usually aim at compensating for nondispersive distortions, such as additive or short convolutive noise.
	\item The arcs in Figures~\ref{fig:vertall}(c) and \ref{fig:vertdet}(b) illustrate that the observed vector $\y_n$ does not only depend on the state $q_n$ (or mixture component $k_n$) through $\x_n$.
		As a consequence, one can deduce that the compensation parameters do depend on the phonetic content, as in Subsections~\mbox{\ref{sec:examples}-\ref{ssec:jud}}, \mbox{\ref{sec:examples}-\ref{ssec:cmllr}}, and \mbox{\ref{sec:examples}-\ref{ssec:mllr}}.
	\item The graphical model representation also succinctly highlights whether a Bayesian modeling paradigm is applied,
		as in Figures~\ref{fig:vertdet}(c) and \ref{fig:Bayes_vs_UncDec}(b), or not,
		as in Figures~\ref{fig:vertdet}(a) and (b).
	\item The existence of the additional latent variable $\x_n$ in most of the presented Bayesian network representations
		expresses that an explicit observation model or an implicit statistical model between
		the clean and the corrupted features is employed.
		In contrast, the graphical representations in Figures~\ref{fig:vertdet}(c) and \ref{fig:HMM_CO_cond} 
		show that -- instead of a distinct compensation step -- a modified \hmm topology is used.
\end{itemize}

In summary, the condensed description of the various concepts from the same Bayesian network perspective
shall offer other researchers the possibility to more easily exploit or combine existing techniques and to link their own algorithms to the presented ones.
This seems all the more important as the recent acoustic modeling approaches based on \dnns raise new challenges for the conventional robustness techniques \cite{li_overview_2014}.
As pointed out in \cite{li_overview_2014}, a possible solution for exploiting the traditional \gmm-based robustness approaches 
-- such as the ones reviewed here -- within the deep learning paradigm
could be the use of \dnn-derived bottleneck features in a \gmm-\hmm, which offers various possibilities for further research.
\ifCLASSOPTIONcaptionsoff
  \newpage
\fi

\bibliographystyle{IEEEtran}
%

%
%
%
%
%
%
%
%
%
%
%
%

%
%
%
%
%
%
%
%
%
%

%
%
%

%
%
%
%

%
%
%

%
%
%
%
%
%
%

%
%
%
%
%

%

%
%
%

%
\end{document}